\documentclass[conference]{IEEEtran}
\IEEEoverridecommandlockouts

\usepackage{hyperref}
\usepackage{cite}
\usepackage{amsmath,amssymb,amsfonts}
\usepackage{algorithmic}
\usepackage{graphicx}
\usepackage{textcomp}
\usepackage{xcolor}
\def\BibTeX{{\rm B\kern-.05em{\sc i\kern-.025em b}\kern-.08em
    T\kern-.1667em\lower.7ex\hbox{E}\kern-.125emX}}

\usepackage{booktabs}
\usepackage{subcaption}
\usepackage{multicol}
\usepackage{multirow}
\usepackage{tabularx}

\usepackage{acronym}
\acrodef{GOOS}[GOOS]{Global Ocean Observing System}
\acrodef{Argo}[Argo]{Array for Real-time Geostrophic Oceanography}
\acrodef{GLOSS}[GLOSS]{Global Sea Level Observing System}
\acrodef{EMSO}[EMSO]{European Multidisciplinary Seafloor Observatory}

\acrodef{QC}[QC]{Quality Control}
\acrodef{DQC}[DQC]{Data Quality Control}
\acrodef{AL}[AL]{Active Learning}
\acrodef{ML}[ML]{Machine Learning}
\acrodef{DL}[DL]{Deep Learning}
\acrodef{MLP}[MLP]{Multi-layer Perceptron}
\acrodef{ODEAL}[ODEAL]{Outlier Detection-Enhanced Active Learning}
\acrodef{ARIMA}[ARIMA]{Autoregressive Integrated Moving Average}

\acrodef{LightGBM}[LightGBM]{Light Gradient Boosting Machine}
\acrodef{XGBoost}[XGBoost]{Extreme Gradient Boosting}
\acrodef{CatBoost}[CatBoost]{Categorical Boosting}
\acrodef{KNN}[KNN]{k-nearest Neighbor}
\acrodef{OCSVM}[OCSVM]{One-Class Support Vector Machine}
\acrodef{iForest}[iForest]{Isolation Forest}
\acrodef{LOF}[LOF]{Local Outlier Factor}
\acrodef{CBLOF}[CBLOF]{Cluster-Based Local Outlier Factor}

\acrodef{Kappa}[Kappa]{Cohen's kappa}
\acrodef{QBC}[QBC]{Query-By-Committee}
\acrodef{RS}[RS]{Random Sampling}
\acrodef{US}[US]{Uncertainty-based Sampling}
\acrodef{CES}[CES]{Consensus-Entropy-based Sampling}

\begin{document}

\title{Ocean Data Quality Assessment through Outlier Detection-enhanced Active Learning}

\author{
\IEEEauthorblockN{Na Li\IEEEauthorrefmark{1}, 
Yiyang Qi\IEEEauthorrefmark{2}\IEEEauthorrefmark{1}, 
Ruyue Xin\IEEEauthorrefmark{1}, 
Zhiming Zhao\IEEEauthorrefmark{1}\IEEEauthorrefmark{3}}

\IEEEauthorrefmark{1}Multiscale Networked System, Informatics Institute, University of Amsterdam, Netherlands\\
\IEEEauthorrefmark{2}Computer Science Department, Vrije Universiteit Amsterdam, Amsterdam, Netherlands\\
\IEEEauthorrefmark{3}LifeWatch ERIC Virtual Lab and Innovation Center (VLIC), Amsterdam, Netherlands\\
n.li@uva.nl, y.qi@student.vu.nl, r.xin@uva.nl, z.zhao@uva.nl
}

\maketitle
\begin{abstract}
Ocean and climate research benefits from global ocean observation initiatives such as Argo, GLOSS, and EMSO. The Argo network, dedicated to ocean profiling, generates a vast volume of observatory data. However, data quality issues from sensor malfunctions and transmission errors necessitate stringent quality assessment. Existing methods, including machine learning, fall short due to limited labeled data and imbalanced datasets. To address these challenges, we propose an \ac{ODEAL} framework for ocean data quality assessment, employing \ac{AL} to reduce human experts' workload in the quality assessment workflow and leveraging outlier detection algorithms for effective model initialization. We also conduct extensive experiments on five large-scale realistic Argo datasets to gain insights into our proposed method, including the effectiveness of \ac{AL} query strategies and the initial set construction approach. The results suggest that our framework enhances quality assessment efficiency by up to 465.5\%  with the uncertainty-based query strategy compared to random sampling and minimizes overall annotation costs by up to 76.9\% using the initial set built with outlier detectors. 
\end{abstract}


\section{Introduction}
To support ocean and climate research, several international ocean observation programs and projects, such as \ac{Argo}~\cite{argo2001argo}, \ac{GLOSS}~\cite{merrifield2009global}, \ac{EMSO}~\cite{favali2009emso}, have been launched for observations and measurements at different sea depths~\cite{lin2020ocean}. The \ac{Argo} observation network, dedicated to profiling the global ocean, comprises thousands of profilers that produce enormous observatory data over time. 
However, the data records usually suffer from quality problems caused by sensor damage, equipment malfunction, and data transmission errors, which may potentially lead to inaccurate scientific conclusions. Thus, it is of great significance to conduct data \acf{QC} before it can be used for any downstream applications~\cite{cummings2011ocean}. Due to the strict requirements on the data credibility, the \ac{QC} process is mainly done by domain experts or involves a high degree of human engagement, and hence is extremely laborious and time-consuming~\cite{abeysirigunawardena2015data}. 

Existing studies have proposed automated and semi-automated data quality assessment approaches to assist \ac{QC} experts. Traditional methods apply a sequence of rule-based statistical tests on the data instances~\cite{abeysirigunawardena2015data,diamant2020cross,skaalvik2023challenges}. These methods rely on pre-defined quality criteria and are subject to specific observation types. As a result, it can only perform basic examination for the validity of data and still depends on manual inspection for accurate outputs. 
An emerging research direction is to exploit \ac{ML} methods for ocean data quality assessment~\cite{zhou2018data,castelao2021machine,mieruch2021salaciaml,demirel2021deep}. One branch of studies utilizes anomaly detection techniques~\cite{xin2023robust} to identify erroneous measurements from the dataset~\cite{zhou2018data,castelao2021machine}. Another line of research employs deep neural networks to classify samples into different quality categories~\cite{mieruch2021salaciaml}.  

Nonetheless, two main challenges remain in automated \ac{QC} research. The first challenge is the lack of labeled data for training \ac{ML} models. Despite the existence of historically labeled data, it is not trivial to transfer knowledge from one dataset to another due to domain shift or concept drift issues. 
Targeting this challenge, we propose utilizing \ac{AL} methods~\cite{settles2009active} to reduce the number of labeled instances required for optimizing \ac{ML} models. \ac{AL} query strategies can select the most informative data instances for labeling and thus improve model performances in a data-efficient manner. In the context of data quality assessment, \ac{AL} can select the most tricky samples that require manual examination and automate the quality assessment for the rest of the samples.

The second challenge is the cold-start problem in \ac{AL} methods posed by severe data imbalance w.r.t. quality labels, with erroneous measurements occupying less than 1\% of the datasets. 
In a typical \ac{AL} scenario, an initial set is required to initialize the classifiers, and the iterative query process follows to refine the classifiers~\cite{settles2009active}. However, a small initial set built from a severely imbalanced dataset likely contains zero erroneous instances, on which\ classifiers are unable to learn meaningful representations for the erroneous instances and consequentially mislead the query process. This is referred to as the cold-start problem. To address this issue, we propose to leverage the outlier detectors for initial set construction, which can increase the probability for initial sets to include erroneous samples and improve their effectiveness in initializing classifiers. 

In this paper, we present an \acf{ODEAL} framework for ocean data quality assessment, which applies \ac{AL} to reduce the workload of \ac{QC} experts in a semi-automated data quality assessment workflow and employs outlier detection algorithms to initialize the learning models effectively with a minimal number of samples. To validate our method, we conducted extensive experiments on five realistic datasets provided by Argo~\cite{argo2000argo}. The results suggest that the \ac{AL} method can increase the F1-score by 465.5\% compared to random sampling, and the outlier detection-initialized approach reduces overall annotation cost by 76.9\%. Our contributions are three-fold: 

\begin{itemize}
    \item presenting a novel \ac{AL}-based data quality assessment framework to reduce the workload of human analysts;  
    \item proposing using outlier detection to construct the initial set for a highly imbalanced dataset to solve the cold-start problem and minimize the overall annotation cost; 
    \item providing empirical evidence and insights for the effectiveness of the proposed method via extensive experiments. 
\end{itemize}
The related datasets and codes can be found here\footnote{\url{https://github.com/QCDIS/odeal}}.

\section{Related work}
Automated and semi-automated \ac{QC} approaches on large-scale ocean observatory data have been studied to support human experts. Classical automated \ac{QC} procedures involve constant value check, spike and step check, range check, and stability check~\cite{abeysirigunawardena2015data,diamant2020cross,skaalvik2023challenges} to screen out gross errors from measurements but are subject to human experts for fine-grained quality examination. More advanced methods utilize \ac{ML} models, such as \ac{MLP} and ARIMA, to analyze large-scale data~\cite{castelao2021machine,zhou2018data,mieruch2021salaciaml}. 
A common approach is using anomaly detection for quality assessment, considering the infrequent occurrence of erroneous samples. Castelão~\cite{castelao2021machine} characterizes the typical behavior of the data by estimating a probability density function (PDF) for each feature, and the survival function (SF) of the estimated PDF is used to quantify the anomalousness of a certain measurement. This approach reduces the error by at least 50\% when applied to 13 years of hydrographic profiles. 
Zhou et al.~\cite{zhou2018data} focus on time-series data and employ the ARIMA model to detect erroneous measurements. The proposed method was applied to pH and CTD data from the Xiaoqushan Seafloor Observatory and achieved an F1 score of 0.9506 in outlier detection. However, ARIMA is a time series forecasting model that requires clean and well-preprocessed time series data for training, which implicitly demands quality labels for historical data points. 

Different from the above methods, Mieruch et al.~\cite{mieruch2021salaciaml} frame data quality assessment as a binary classification problem, where each sample is categorized as of \emph{good} or \emph{bad} qualities. They present SalaciaML, aiming to mimic the skillful \ac{QC} experts with a deep learning artificial neural network in identifying potentially erroneous samples. A \ac{MLP} is trained on more than 2 million temperature measurements over the Mediterranean Seas spanning the last 100 years and detects correctly more than 90\% of all good and/or bad data in 11 out of 16 Mediterranean regions. However, it does not consider the correlations between different water properties. 

The high accuracy achieved by the reviewed studies is at the cost of a large amount of labeled data, which is usually the most tricky problem in real-world scenarios. Therefore, our goal is to decrease the demand for labeled data in data quality assessment by leveraging \ac{AL} methods.

\section{Methodology}
\subsection{Problem Definition}

To formally define the problem, let $x^{s}_t\in \mathbb{R}$ be an observation record measured at time point $t$ by sensor $s$. Given a time frame $T$, data records produced by a set of sensors $S$ are $\{[x^{s_i}_t, x^{s_2}_t, \cdots, x^{s_d}_t]\}\in \mathbb{R}^{|T|\times |S|}, s_i\in S, t\in T$. For simplicity, we will denote $[x^{s_i}_t, x^{s_2}_t, \cdots, x^{s_d}_t]$ as $x_t$ thereafter. The task of quality assessment is to assign a quality flag $\hat y_t$ to a data instance $x_t$. 
In this work, we only consider binary quality labels, where each instance is classified as \emph{with} OR \emph{without quality issues}, and thus we have $\hat y_t \in \{0, 1\}$. The quality flag $1$ means there are quality issues in the data instance, while $0$ means there is no quality problem. 






\subsection{\acl{ODEAL} Framework}
\begin{figure*}[h!]
    \centering
    \includegraphics[width=0.7\textwidth]{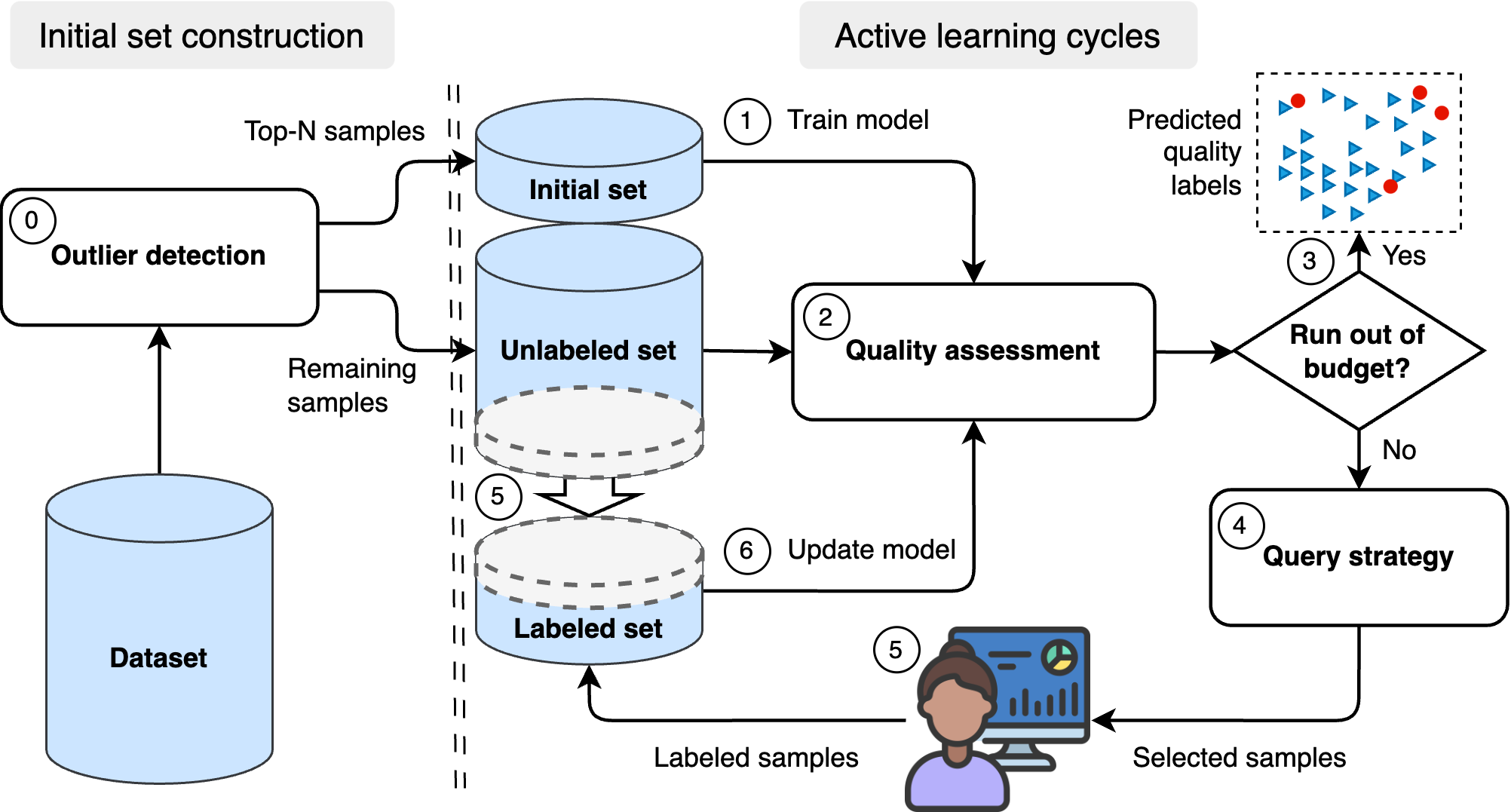}
    \caption{High-level structure of the proposed \ac{ODEAL} framework}
    \label{fig:high_level_framework}
\end{figure*}

We propose an \ac{AL}-based data quality assessment framework called \acf{ODEAL}, which aims to label all the data instances with high accuracy while minimizing the annotation cost, i.e., the number of samples that need to be manually labeled. It operates under the pool-based \ac{AL} paradigm. Figure~\ref{fig:high_level_framework} depicts the high-level structure of the proposed framework. It consists of the \emph{Initial set construction} and the \emph{Active learning cycles} phases. In the first phase, we build an initial set $D_I$ with anomaly detectors, which select the top-$N_I$ anomalous samples from the target dataset $D$, where $N_I$ is the size of $D_I$. And the remaining samples are put into the unlabeled set $D_U$. $D_I$ is sent to human experts for labels and then used for classifier initialization. 

During the \ac{AL} cycles, the labeled set $D_L$ stores all the labeled instances generated by the human annotators. The target models are quality assessors that assign quality labels to data instances. A human expert is involved as the oracle providing legitimate quality labels for queried samples. 
The \ac{AL} workflow is described below: 
\begin{enumerate}
    \item The quality assessment models are trained on the initial set (Step 1). 
    \item The quality assessment models predict the quality labels on all instances in the unlabeled set (Step 2). 
    \item If the budget is exhausted or the models are confident enough for the predicted results, output the predictions (Step 3); 
    \item Otherwise, the models produce intermediate results for the query strategy to select the $K$ most informative instances (Step 4). 
    \item The selected instances are presented to human annotators for labeling. Meanwhile, they are moved from the unlabeled set to the labeled set (Step 5). 
    \item The quality assessment models are updated using the labeled set (Step 6).  
    \item Repeat Steps 2, 4, 5, and 6 until the annotation budget exhausts or the confidence threshold is reached. 
\end{enumerate}

For implementation, the annotation budget and the confidence threshold should be defined under the operational condition depending on the requirements. It is also application-dependent in terms of quality assessment model selection and query strategy development. We illustrate our choices of quality assessment models and query strategy in the ocean observation use case context. However, it is at the readers' discretion to make their decisions. 

\subsection{Initial Set Construction}

The importance of the initial set is often overlooked by previous studies. Most of them create an initial set using a pre-defined number of randomly selected samples and exclude the labeling of the initial set from the annotation budget. There are two problems with it. First, in real applications, the labels of the initial set are also provided by human experts and should be included in the overall budget. Second, the initial set can significantly affect the following \ac{AL} process. Insufficient initial instances may cause cold-start issues, while redundant initial instances will result in a wastage of annotation costs. Moreover, some classifiers, e.g., CatBoost, can only learn from different classes of data, which will require a large initial set when the abnormal samples are elusive in the dataset, i.e., a highly imbalanced dataset. 

To address the above challenges, we propose exploiting outlier detectors (or anomaly detectors) for forming the initial set $D_I$. The objective here is to identify anomalies for inclusion in $D_I$, while maintaining its compact size. Because there is a scarcity of labeled data, the outlier detection algorithms must function in an unsupervised manner or be fine-tuned with a minimal amount of labeled samples. While it is possible to utilize the outlier detectors alone for implementing \ac{AL} methods, updating an unsupervised outlier detector with labeled data is complex. As a result, our research places its emphasis on employing outlier detectors to only build the initial set. 
Let $\psi$ be an outlier detector, and then we have an outlier score $\mu_t=\psi(x_t)$ for data instance $x_t$. $D_I$ is composed of the top-$N_I$ samples from the dataset $D$ ranked by outlier scores, i.e., 
\begin{equation}
    D_I = \{x_t \mid \mu_t \in \text{top-$N_I$ outlier scores}\}, 
\end{equation}
where $N_I$ is the size of $D_I$. This method increases the possibility of including abnormal samples in a small initial set to warm up the classifiers and is expected to be effective in severely imbalanced datasets. 
In this work, we consider three common outlier detection algorithms, i.e., \ac{iForest}~\cite{liu2008isolation}, \ac{OCSVM}~\cite{scholkopf1999support}, and \ac{LOF}~\cite{breunig2000lof}.

\subsection{Quality Assessment Model Selection}
The classifiers (also called target models or learners in the \ac{AL} paradigm) being investigated in our study include \ac{KNN}~\cite{fix1989discriminatory}, \ac{XGBoost}~\cite{chen2016xgboost}, \ac{CatBoost}~\cite{prokhorenkova2018catboost} and \ac{LightGBM}~\cite{ke2017lightgbm}. 
\ac{KNN} is an instance-based algorithm that classifies data points by considering the class of their nearest neighbors. \ac{XGBoost} is an ensemble algorithm utilizing gradient boosting, excelling in structured data tasks, and featuring sequential model correction. \ac{CatBoost} is a gradient boosting approach tailored for categorical data, automatically handling categorical features during training. \ac{LightGBM} is a high-efficiency gradient boosting framework using histogram-based techniques for faster training, suitable for large datasets. \ac{XGBoost}, \ac{CatBoost}, and \ac{LightGBM} are effective in dealing with imbalanced data, as they assign higher weights to misclassified samples, which give more attention to the minority class.


\subsection{Query Strategy}
\label{sec:query_strategy}


We adopt the \ac{US} strategies for classifiers to select instances. These query strategies are based on the hypothesis that the more uncertain one classifier is about the predictive result, the more informative the data sample is. Thus, it will be more useful for optimizing the classifier. There are various ways to compute the \emph{uncertainty} for predictions, such as \emph{confidence}, \emph{margin}, and \emph{entropy}. In binary classification scenarios, the corresponding query strategies, i.e., \emph{least confident}, \emph{smallest margin}, and \emph{maximum entropy}, become equivalent~\cite{settles2009active}. Here we describe the prediction entropy query strategy. Supposing that a classifier $\phi$ can output the class probability $P_\phi(\hat y_t|x_t), \hat y_t \in \{0,1\}$ for a sample $x_t$, the entropy-based uncertainty of the classifier is defined as: 
\begin{equation}
    H_{EN}(x_t) = - \sum_{\hat y_t \in \{0,1\}}P_\phi(\hat y_t|x_t)\log P_\phi(\hat y_t|x_t). 
\end{equation}
The sample with maximum entropy will be requested for labeling: 
\begin{equation}
    x_{EN}^\ast = \arg \max_t H_{EN}(x_t)
\end{equation}


\section{Experiments}


\subsection{Experiment Setup}

\subsubsection{Dataset}
We use ocean observatory data provided by Argo~\cite{argo2000argo}, an international program that collects subsurface ocean water properties such as temperature, salinity, and currents across the global earth using a fleet of robotic instruments. The instruments called \emph{floats} or \emph{profilers}, drift with the ocean currents and move up and down between the surface and a mid-water level. 
We build five datasets using records produced by five similar floats in order to reduce the influence of environmental and operational factors. The five floats were equipped with the same sensors and deployed in the same month (March 2019) within the same area (Atlantic Ocean). 
We consider totally six features, including \emph{datetime}, \emph{latitude}, \emph{longitude}, \emph{pressure}, \emph{temperature} and \emph{salinity}. 
The datasets have extensive quality labels for all the data samples provided by human analysts. To mimic the real usage scenario, we only use labels of queried samples, treating them as acquired from human experts. When processing the quality flags, we treat samples labeled as \emph{good data} as error-free, denoted as 0, and others as erroneous data, denoted as 1. Originally, the quality flags were assigned for each feature. To get a global quality label for the data, we treat the data instance as 0 only if all the features have the label 0. 

The statistics of the datasets are listed in Table~\ref{tab:dataset_statistics}. The dataset $DS_{high}$ exhibits a substantial error rate of 33.72\%, whereas the remaining four datasets, namely $DS_{low}$1-4, demonstrate exceedingly minimal error rates, all below 1\%.
Each dataset is randomly split into 60\% training, 20\% validation, and 20\% test subsets while maintaining the same error rate for each subset. All the features are first normalized using the Z-score method before feeding into the classifiers.

\begin{table}[h!]
    \centering
    \caption{Statistics of the datasets. }
    \label{tab:dataset_statistics}
    \begin{tabularx}{\columnwidth}{XXXXXX}
    \toprule
      \textbf{Dataset} & \textbf{Float code} & \textbf{Launch date} & \textbf{Error rate} & \textbf{Training samples} & \textbf{Test samples} \\    \midrule
        $DS_{high}$ & 4903217 & 21/03/2019 & 33.72\% & 179,539 & 59,847 \\
        $DS_{low}1$ & 4903218 & 10/03/2019 & 0.84\% & 175,583 & 58,528 \\ 
        $DS_{low}2$ & 4903220 & 07/03/2019 & 0.16\% & 181,009 & 60,337 \\
        $DS_{low}3$ & 4903052 & 20/03/2019 & 0.69\% & 179,000 & 59,667 \\
        $DS_{low}4$ & 4903054 & 23/03/2019 & 0.23\% & 177,455 & 59,152 \\
    \bottomrule
    \end{tabularx}
\end{table}

\subsubsection{Evaluation metrics}
We use \emph{F1-score} to evaluate the performance of the classifiers.

\subsection{Effectiveness of \ac{AL} Query Strategies}
\label{sec:al_query_strategy}

\begin{figure*}
    \centering
    \begin{subfigure}{0.25\textwidth}
        \centering
        \includegraphics[width=\linewidth]{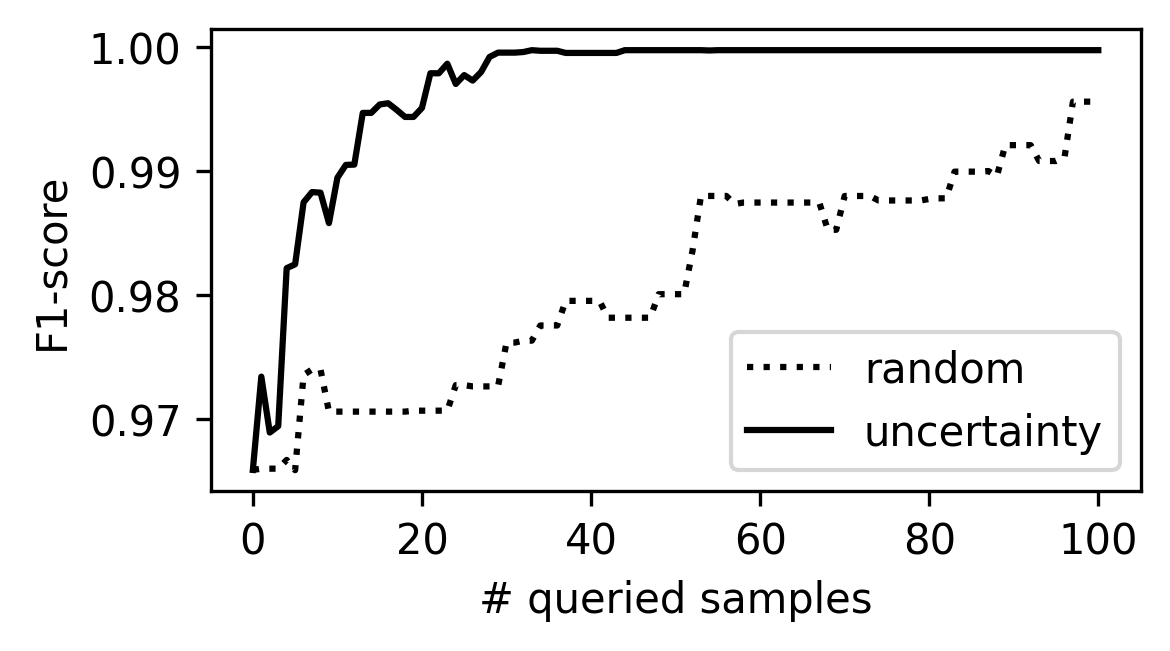}
        \caption{KNN}
    \end{subfigure}%
    \begin{subfigure}{0.25\textwidth}
        \centering
        \includegraphics[width=\linewidth]{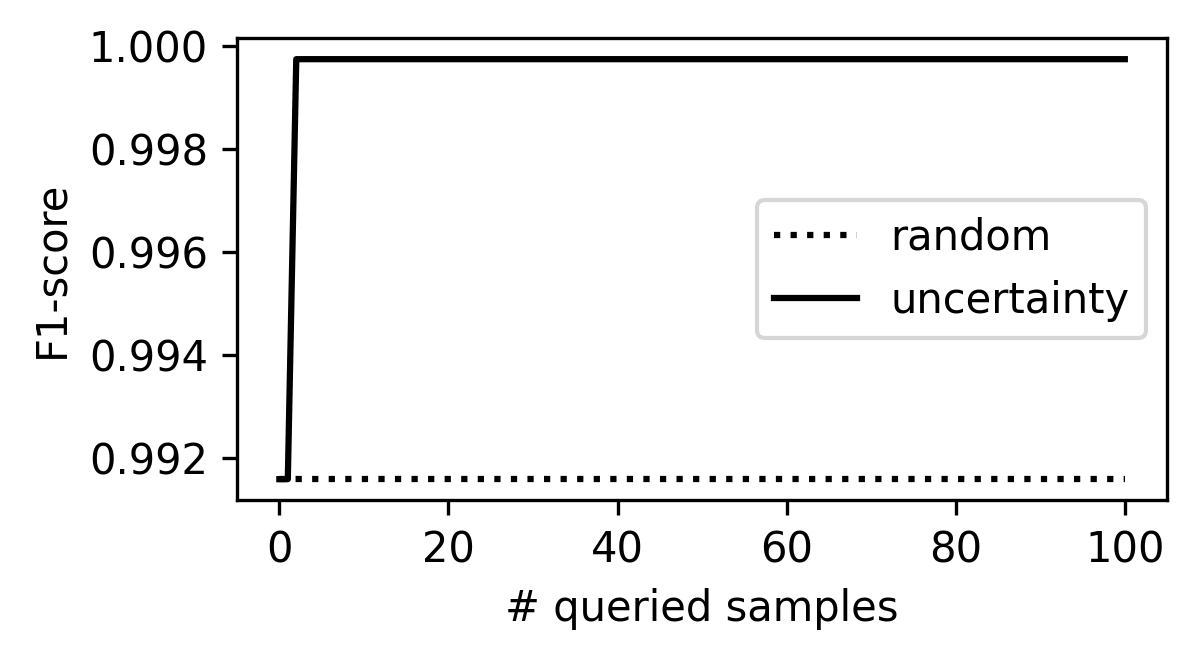}
        \caption{XGBoost}
    \end{subfigure}%
    \begin{subfigure}{0.25\textwidth}
        \centering
        \includegraphics[width=\linewidth]{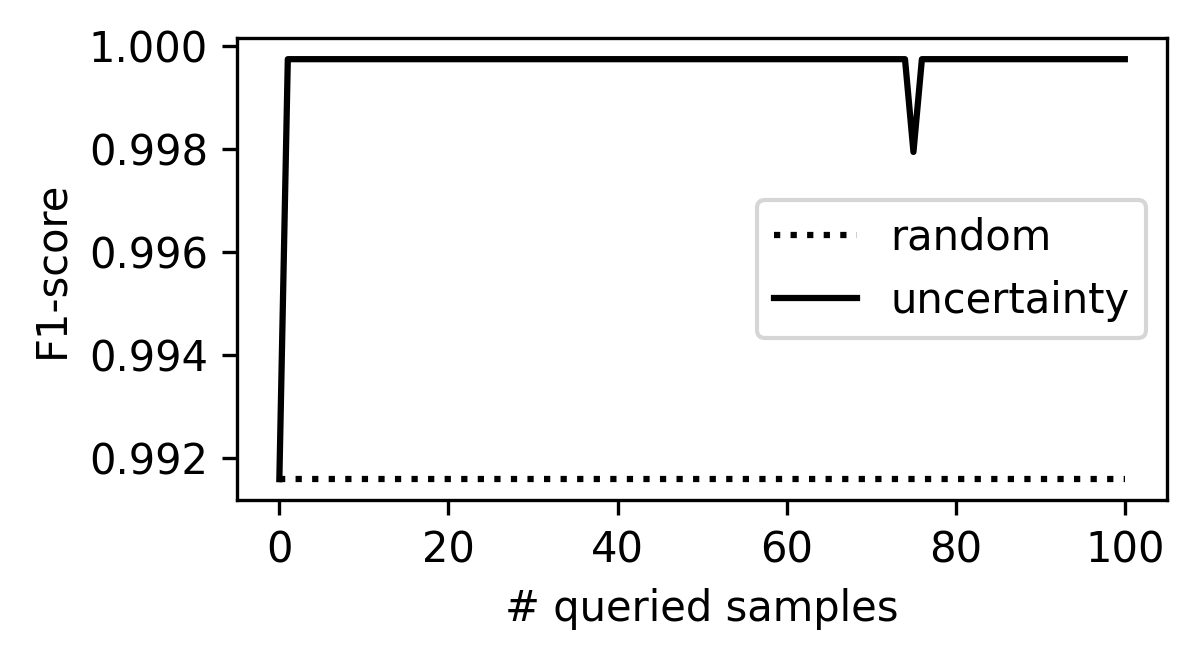}
        \caption{CatBoost}
    \end{subfigure}%
    \begin{subfigure}{0.25\textwidth}
        \centering
        \includegraphics[width=\linewidth]{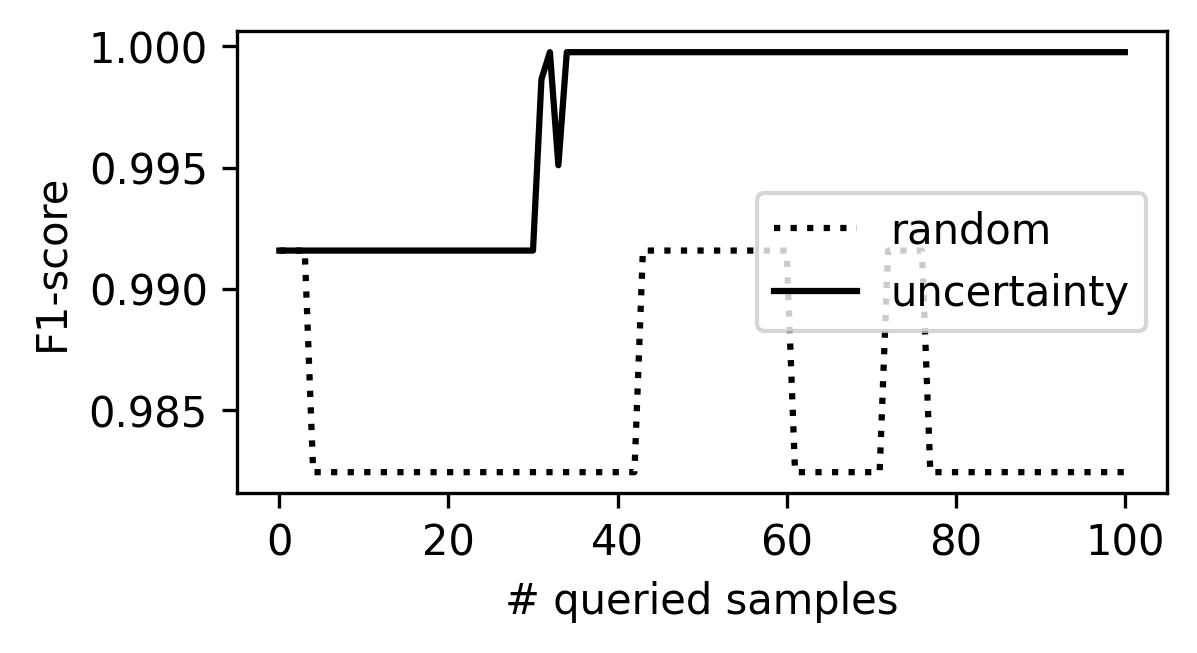}
        \caption{LightGBM}
    \end{subfigure}

    \begin{subfigure}{0.25\textwidth}
        \centering
        \includegraphics[width=\linewidth]{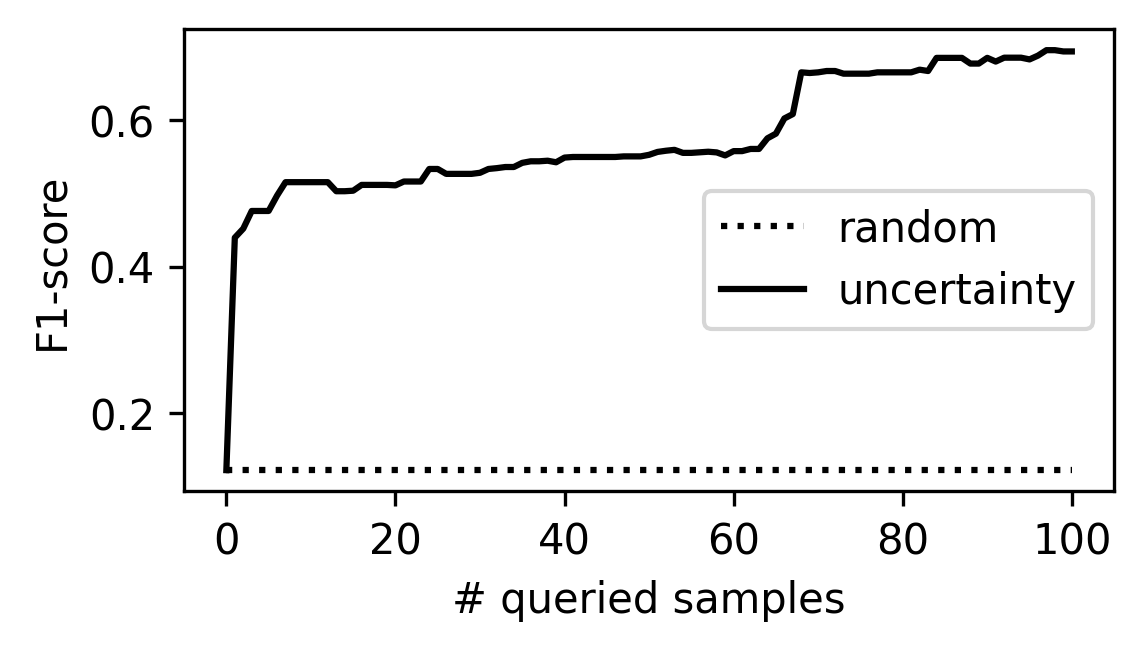}
        \caption{KNN}
    \end{subfigure}%
    \begin{subfigure}{0.25\textwidth}
        \centering
        \includegraphics[width=\linewidth]{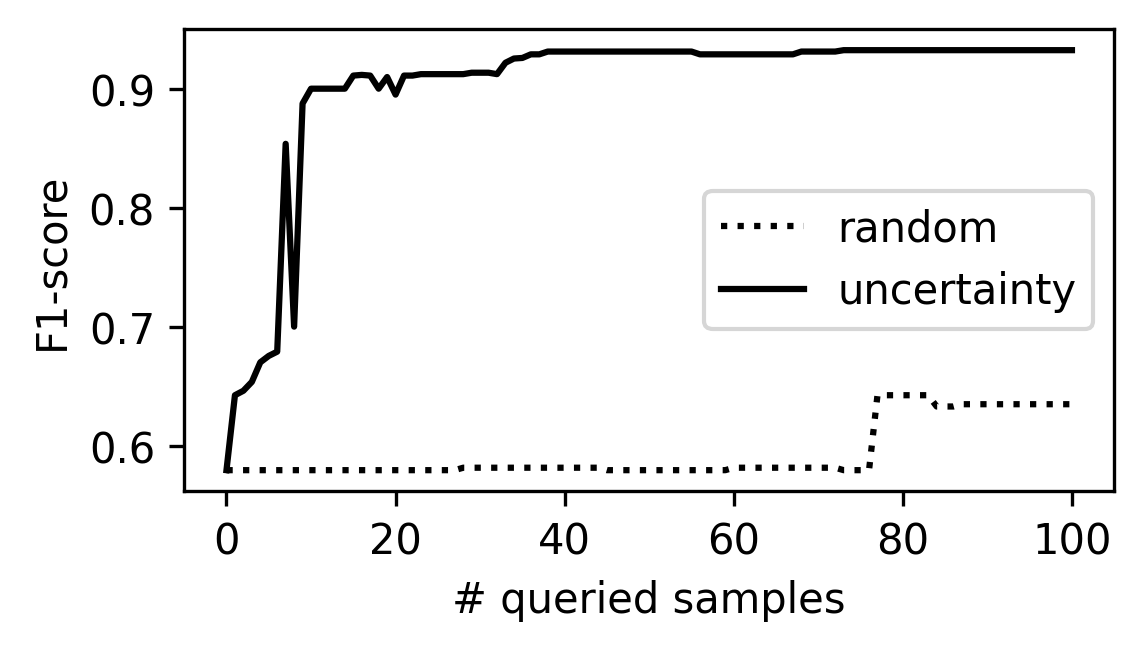}
        \caption{XGBoost}
    \end{subfigure}%
    \begin{subfigure}{0.25\textwidth}
        \centering
        \includegraphics[width=\linewidth]{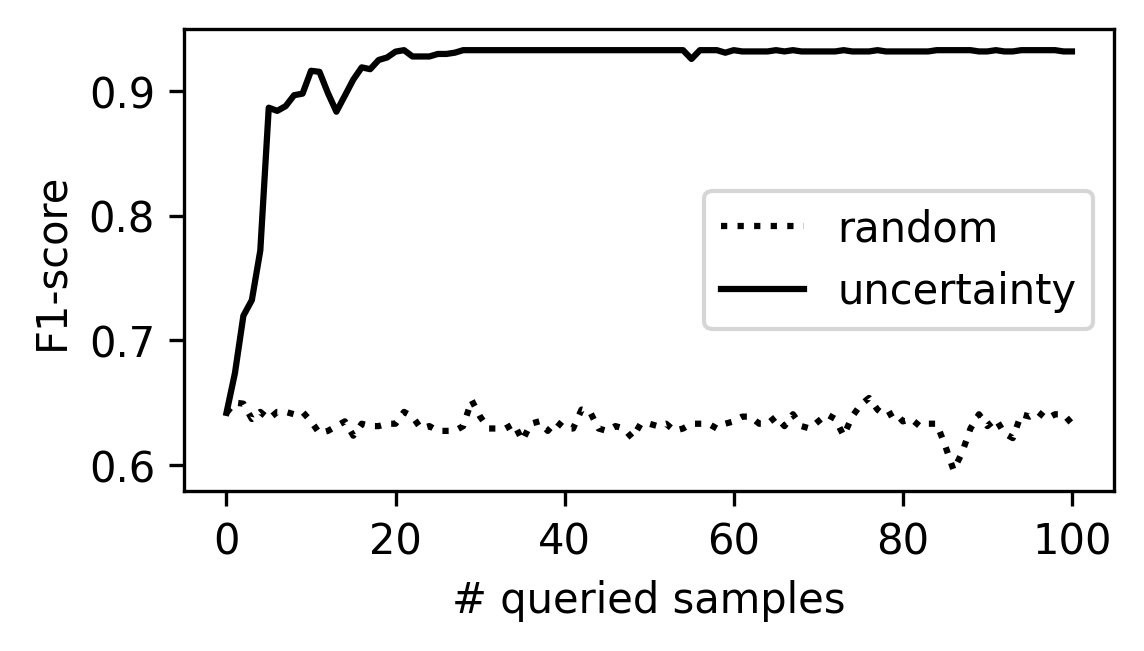}
        \caption{CatBoost}
    \end{subfigure}%
    \begin{subfigure}{0.25\textwidth}
        \centering
        \includegraphics[width=\linewidth]{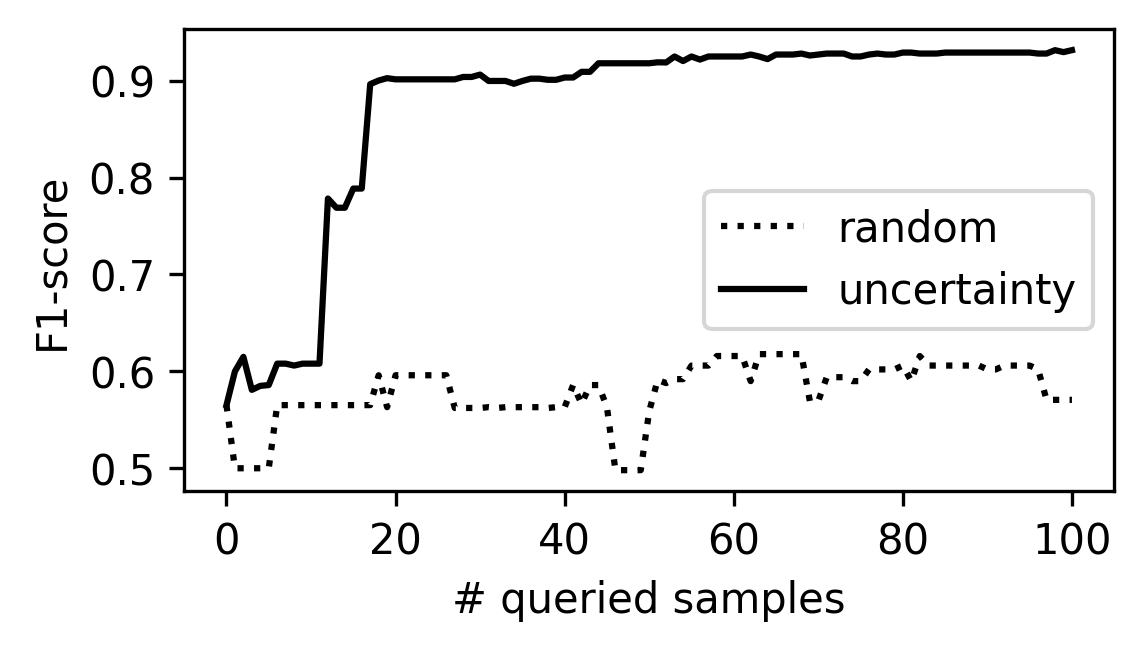}
        \caption{LightGBM}
    \end{subfigure}

    \begin{subfigure}{0.25\textwidth}
        \centering
        \includegraphics[width=\linewidth]{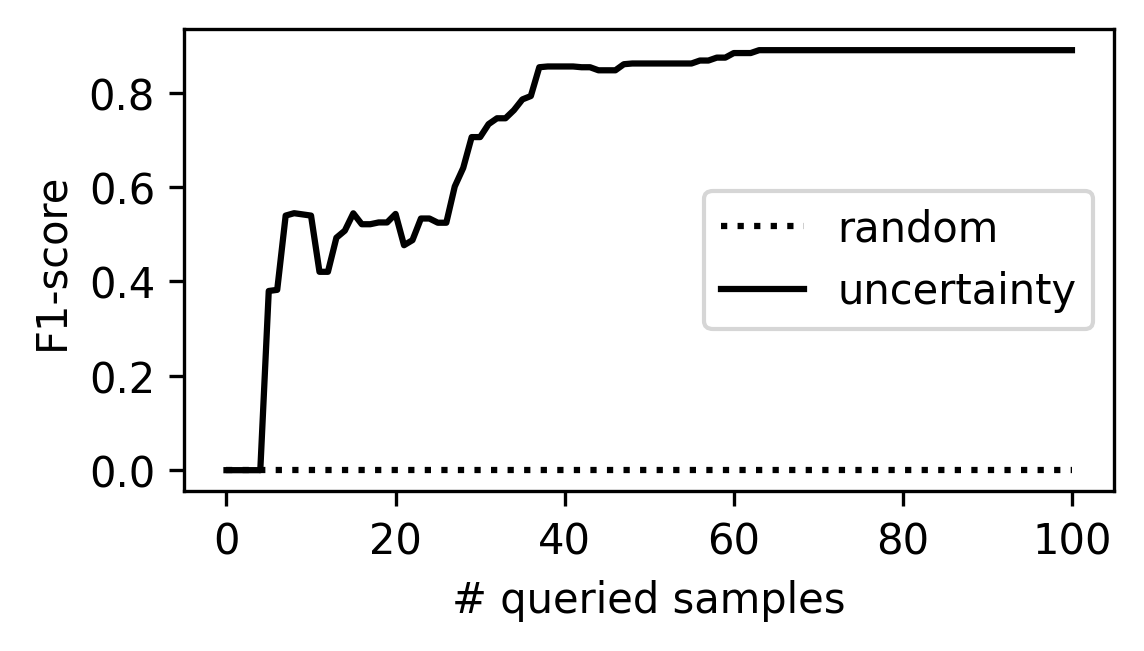}
        \caption{KNN}
    \end{subfigure}%
    \begin{subfigure}{0.25\textwidth}
        \centering
        \includegraphics[width=\linewidth]{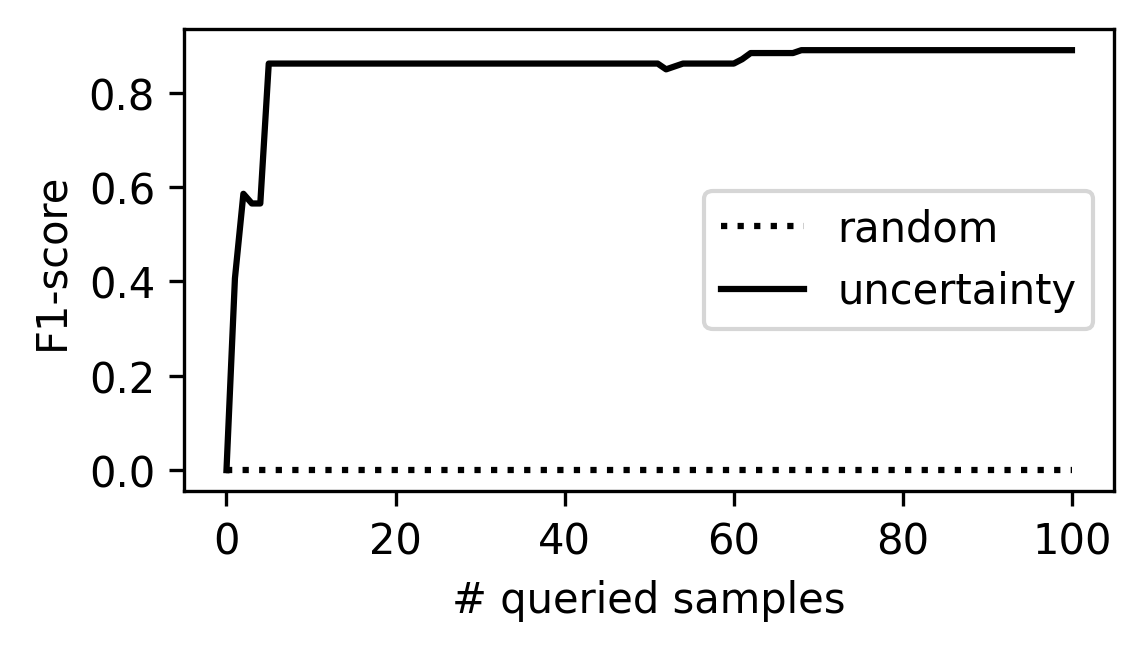}
        \caption{XGBoost}
    \end{subfigure}%
    \begin{subfigure}{0.25\textwidth}
        \centering
        \includegraphics[width=\linewidth]{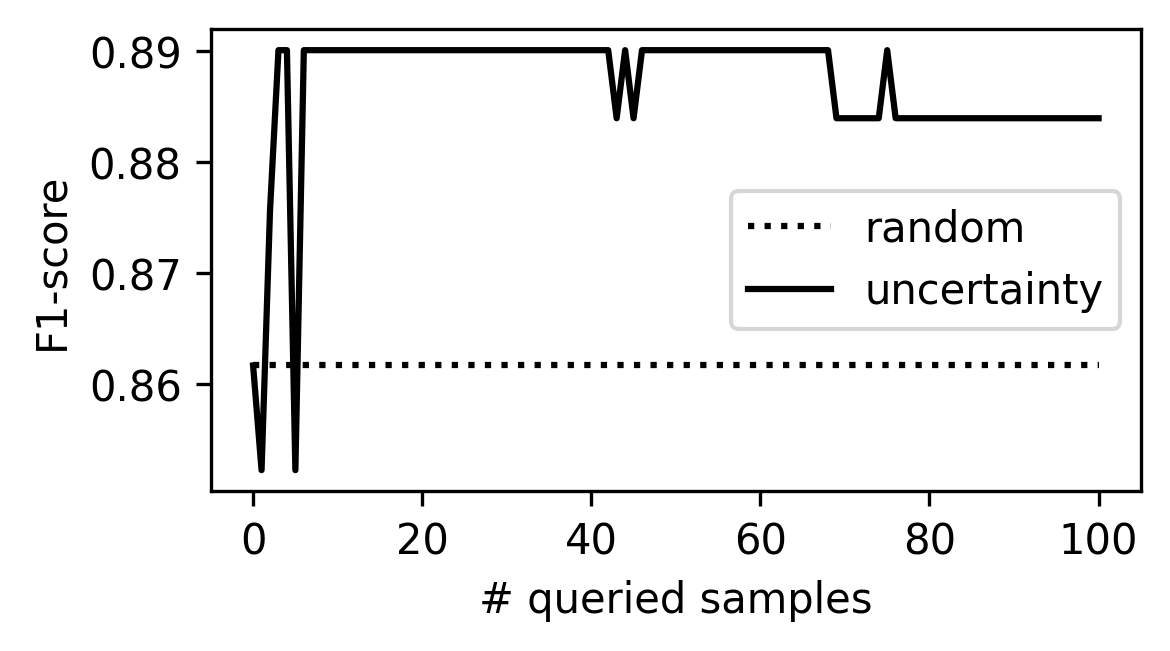}
        \caption{CatBoost}
    \end{subfigure}%
    \begin{subfigure}{0.25\textwidth}
        \centering
        \includegraphics[width=\linewidth]{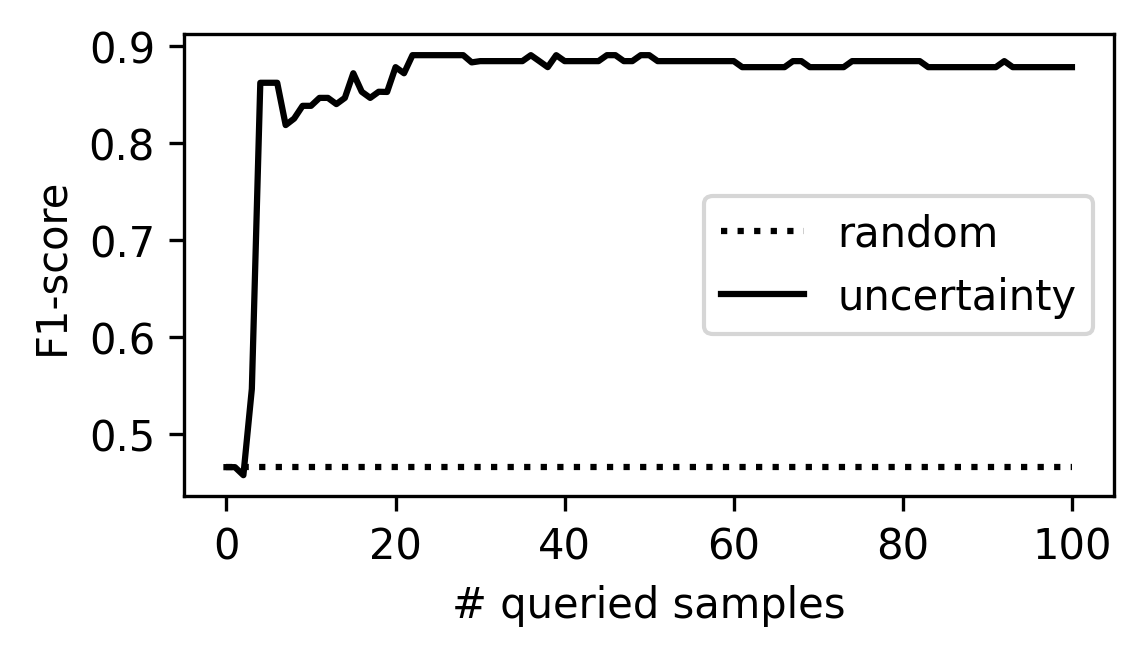}
        \caption{LightGBM}
    \end{subfigure}

    \begin{subfigure}{0.25\textwidth}
        \centering
        \includegraphics[width=\linewidth]{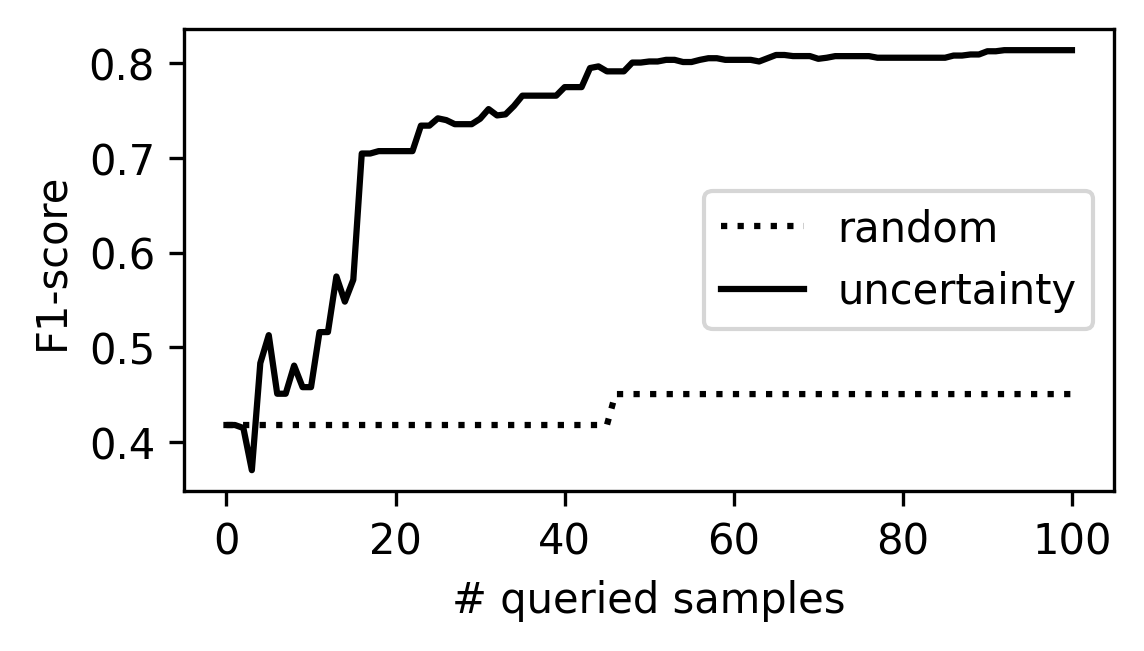}
        \caption{KNN}
    \end{subfigure}%
    \begin{subfigure}{0.25\textwidth}
        \centering
        \includegraphics[width=\linewidth]{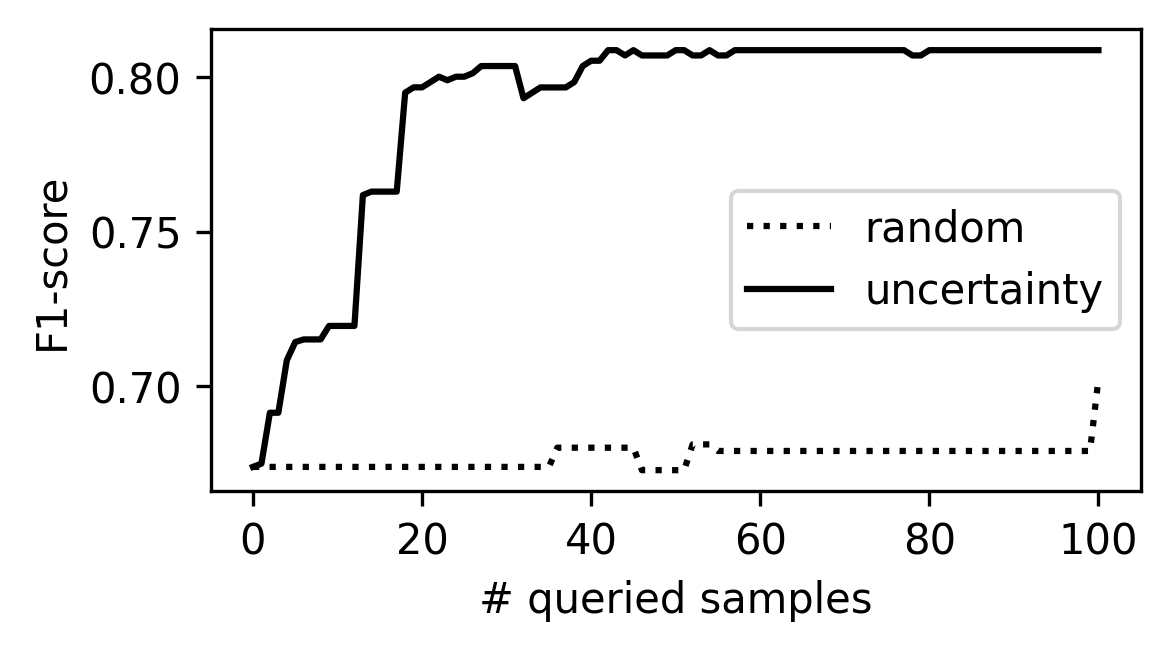}
        \caption{XGBoost}
    \end{subfigure}%
    \begin{subfigure}{0.25\textwidth}
        \centering
        \includegraphics[width=\linewidth]{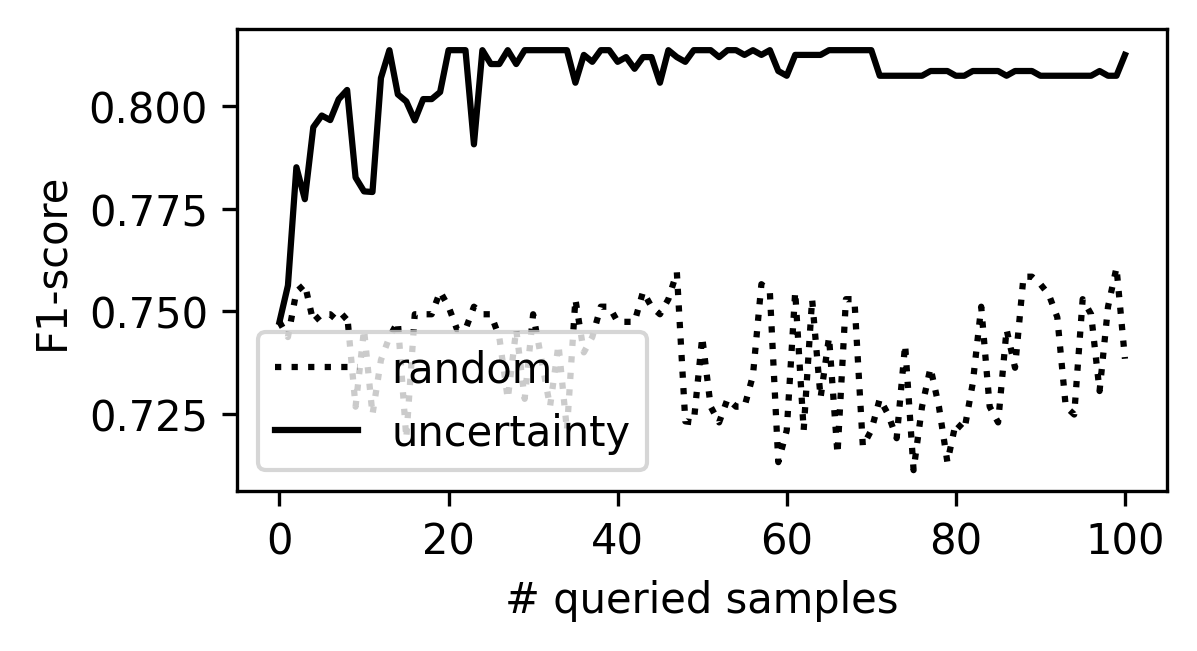}
        \caption{CatBoost}
    \end{subfigure}%
    \begin{subfigure}{0.25\textwidth}
        \centering
        \includegraphics[width=\linewidth]{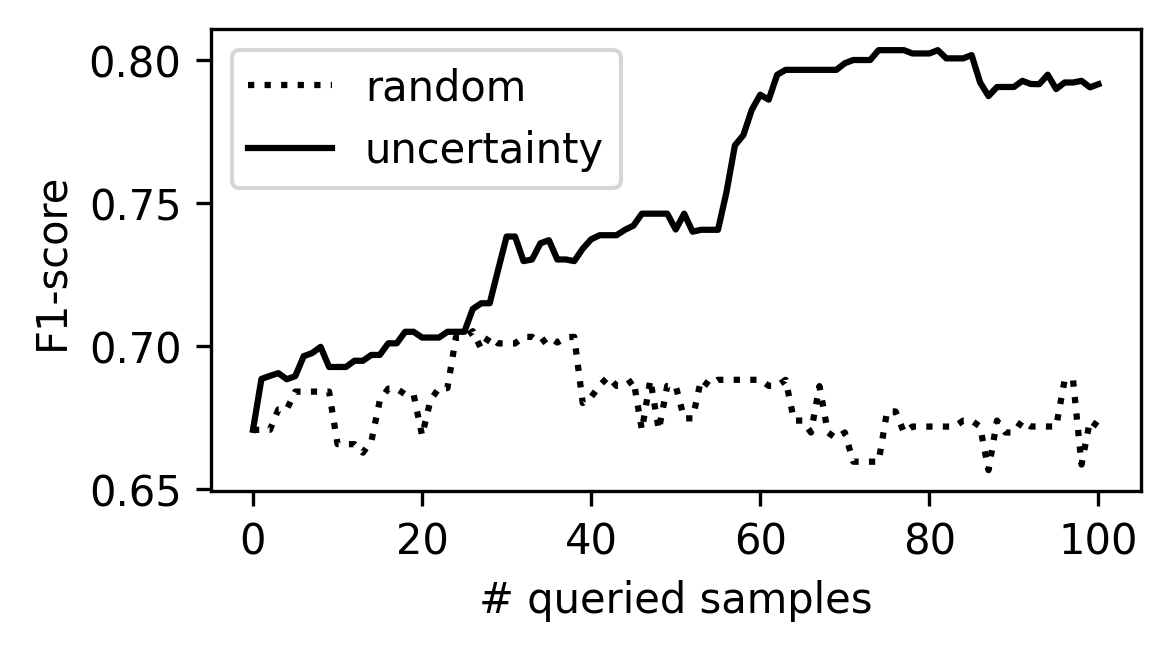}
        \caption{LightGBM}
    \end{subfigure}

    \begin{subfigure}{0.25\textwidth}
        \centering
        \includegraphics[width=\linewidth]{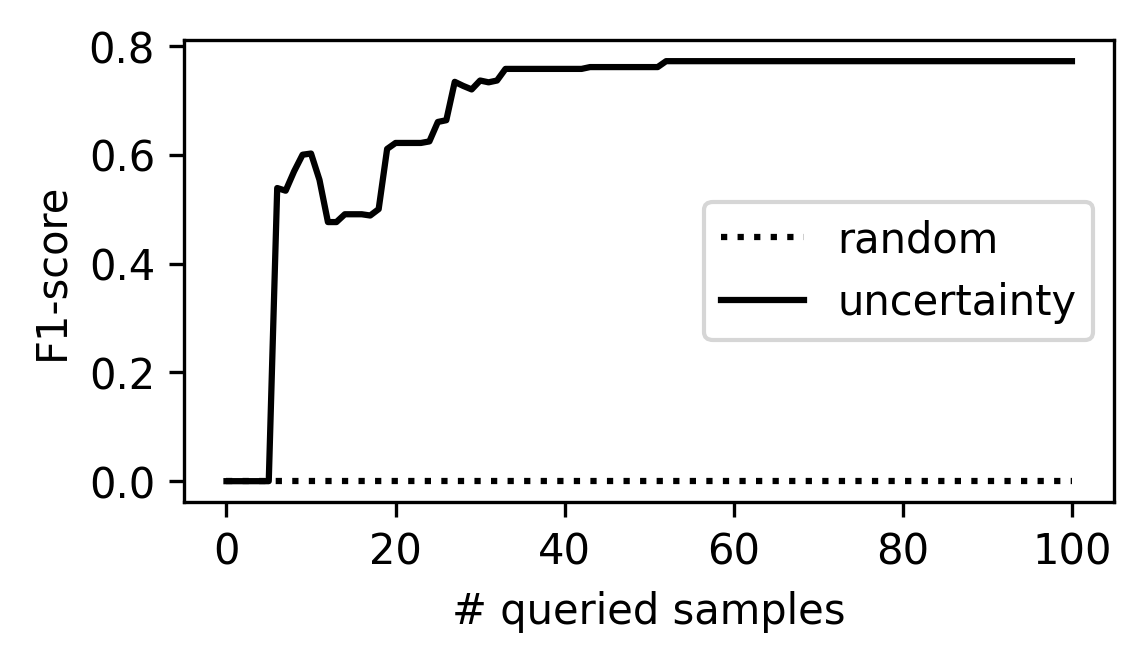}
        \caption{KNN}
    \end{subfigure}%
    \begin{subfigure}{0.25\textwidth}
        \centering
        \includegraphics[width=\linewidth]{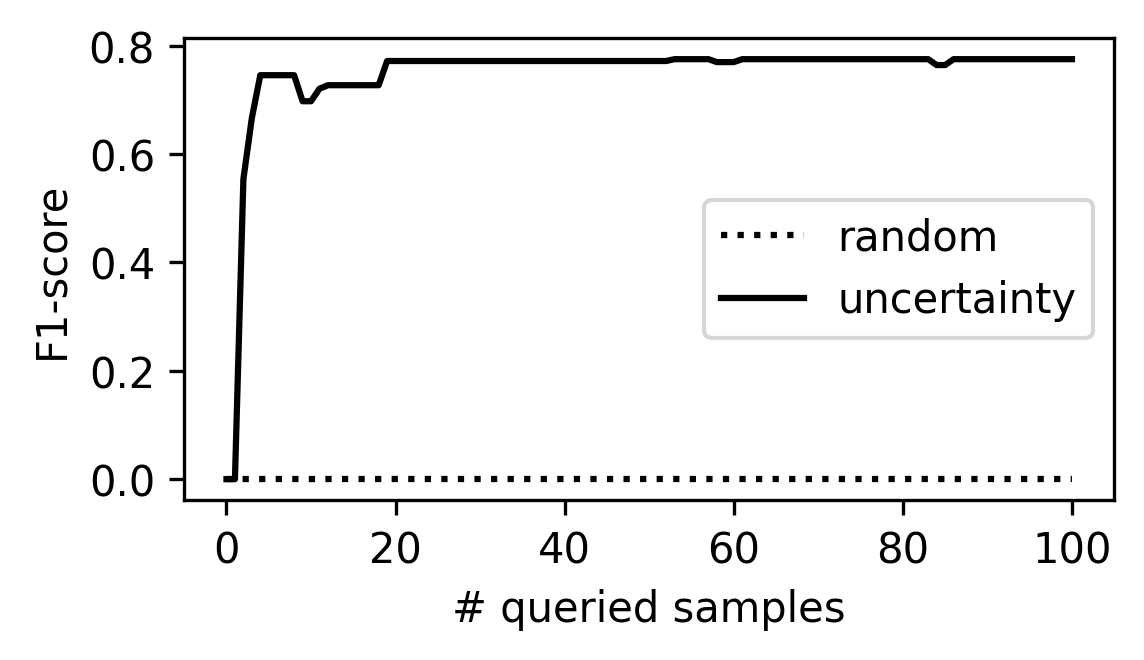}
        \caption{XGBoost}
    \end{subfigure}%
    \begin{subfigure}{0.25\textwidth}
        \centering
        \includegraphics[width=\linewidth]{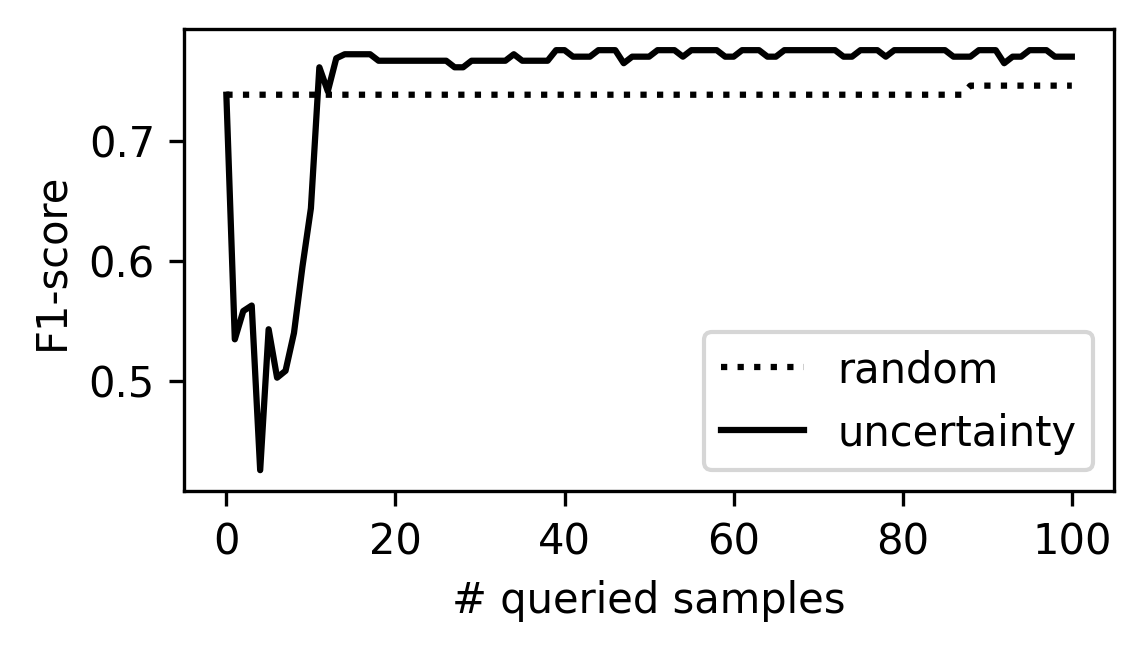}
        \caption{CatBoost}
    \end{subfigure}%
    \begin{subfigure}{0.25\textwidth}
        \centering
        \includegraphics[width=\linewidth]{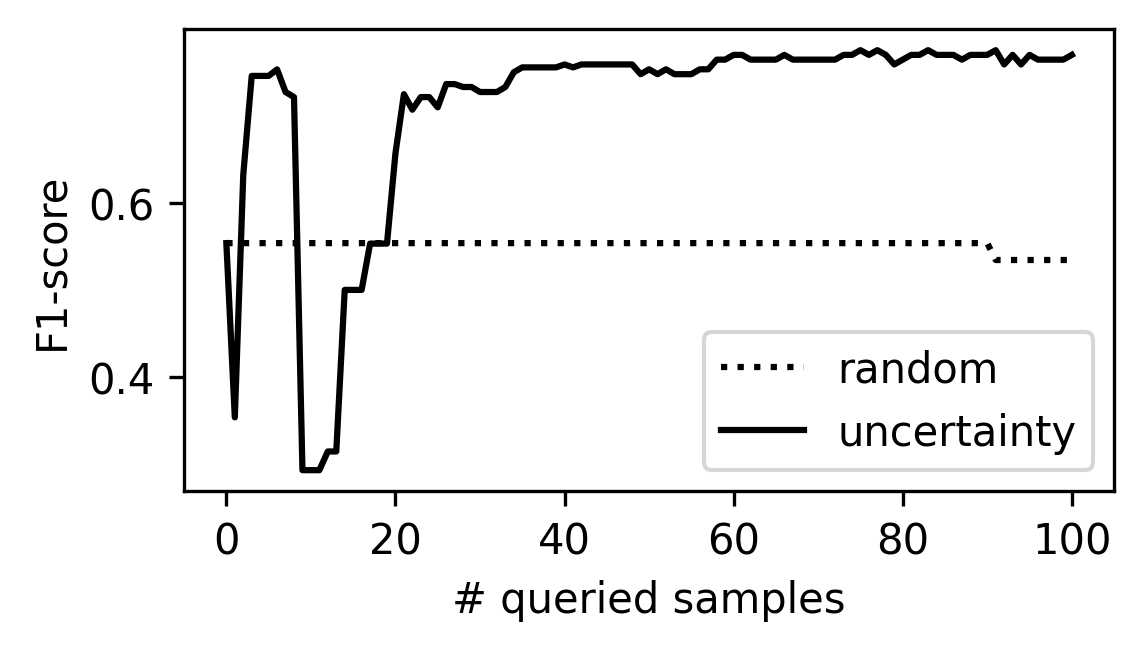}
        \caption{LightGBM}
    \end{subfigure}

    \caption{Comparisons of query strategies on different datasets. Legends denote the query strategies: ``random'' stands for random sampling, and ``uncertainty'' uncertainty-based sampling. (a)(b)(c)(d): $DS_{high}$; (e)(f)(g)(h): $DS_{low}1$; (i)(j)(k)(l): $DS_{low}2$; (m)(m)(o)(p): $DS_{low}3$; (q)(r)(s)(t): $DS_{low}4$. K=1 for all datasets.}
    \label{fig:uncertainty_over_random}
\end{figure*}



We examine the effectiveness of the \acf{US} method by comparing it to \ac{RS}. The initial set is formed by randomly selected instances. Considering the error rates of the datasets, we set the initial set size to 1000 for $DS_{high}$ and 100 for the rest datasets. Figure~\ref{fig:uncertainty_over_random} shows the results for four classifiers on five datasets.  

The results suggest that the \ac{US} method is effective for all datasets, either with high error rates or low error rates, but is remarkably successful on severely imbalanced datasets. 
First, we observe that on $DS_{high}$ \ac{XGBoost} and \ac{CatBoost} rapidly reach their best F1-score with less than 10 queried samples using \ac{US}, while the \ac{RS} method makes no difference in the models' performance throughout the \ac{AL} cycles. \ac{LightGBM} shows similar behavior though it needs more samples to optimize. An exception is \ac{KNN}, where the \ac{RS} method still increases its accuracy but takes much longer to achieve optimal performance compared to the \ac{US} method. Compared with \ac{RS}, \ac{US} improve the F1-score for \ac{KNN}, \ac{XGBoost}, \ac{CatBoost} and \ac{LightGBM} on $DS_{high}$ by 0.4\%, 0.8\%, 0.8\%, and 1.8\%, respectively. The subtle progress is due to the fair performance achieved by classifiers before the \ac{AL} query process. 
Second, \ac{US} completely beats \ac{RS} on four low-error datasets, i.e., $DS_{low}$1-4, regardless of classifiers. On $DS_{low}1$, compared with \ac{RS}, the \ac{US} approach increases the F1-score of \ac{KNN}, \ac{XGBoost}, \ac{CatBoost} and \ac{LightGBM} by 465.5\%, 46.8\%, 47.1\%, and 63.2\%, respectively with merely 100 queried samples. And on $DS_{low}3$, the improvement of F1-score for \ac{KNN}, \ac{XGBoost}, \ac{CatBoost} and \ac{LightGBM} are 80.5\%, 15.1\%, 10.0\%, and 17.8\%, respectively. On $DS_{low}2$ and $DS_{low}4$, \ac{KNN} and \ac{XGBoost} report F1-score of 0 using \ac{RS} but rapidly reach the highest F1-score with \ac{US}. \ac{CatBoost} and \ac{LightGBM} see performance increase of 2.6\% and 88.3\% respectively on $DS_{low}2$, 3.2\% and 44.1\% respectively on $DS_{low}4$, when comparing \ac{US} to \ac{RS}.

\subsection{Effectiveness of using outlier detectors for initial set construction}
We compare \ac{iForest}~\cite{liu2008isolation}, \ac{OCSVM}~\cite{scholkopf1999support}, and \ac{LOF}~\cite{breunig2000lof} to detect outliers on $DS_{low}4$ and report the results in Table~\ref{tab:anomaly_detection}. 
We use \ac{OCSVM} with non-linear kernels (RBF). $N_I/2$ randomly selected samples are used to train \ac{OCSVM}, together with the top-$N_I/2$ predicted anomalous samples to form the initial set. The number of estimators in \ac{iForest} is set to 100. The number of neighbors in \ac{LOF} is set to 10. The contamination value for \ac{iForest} and \ac{LOF} are defined as the dataset's error rate. 

\begin{table}[h!]
    \centering
    \caption{Number of anomalies identified by outlier detection methods on $DS_{low}4$. $N_I$ stands for the initial set size. }
    \label{tab:anomaly_detection}
    \begin{tabular}{lcccc}
    \toprule
    \multirow{2}{*}{\textbf{Method}} & \multicolumn{4}{c}{\textbf{\# anomalies@$N_I$}} \\
    \cmidrule{2-5}
    & \textbf{$N_I$=100} & \textbf{$N_I$=200} & \textbf{$N_I$=300} & \textbf{$N_I$=400} \\
    \midrule
    iForest & 0 & 0 & 0 & 0 \\
    OCSVM & 0 & 0 & 0 & 0 \\
    LOF & 4 & 6 & 14 & 21 \\
    \bottomrule
    \end{tabular}
\end{table}

Compared with using random selection to build the initial set, \ac{LOF} can largely reduce the size of the initial set $D_I$ to contain erroneous instances. According to Table~\ref{tab:anomaly_detection}, \ac{LOF} successfully recognizes 4, 6, 14, and 21 true anomalies within the top-100, 200, 300, and 400 predicted anomalies, respectively, while \ac{iForest} and \ac{OCSVM} fail to identify any anomalies in the same settings. \ac{LOF} is $4/100/0.0023\approx17$ times efficient in building the initial set containing erroneous samples in comparison with random selection. 

Therefore, we exploit \ac{LOF} to construct the initial set $D_I$, denoted as $D_I^{LOF}$, differing from the one formed by randomly selected instances, denoted as $D_I^{RD}$. 
To quantify the efficacy of the LOF-initialized \ac{AL} method, we employ the reduced annotation cost as the evaluation metric. Let $N_I$ and $N_L$ be the numbers of instances in the initial set and in the labeled set, respectively, and then we have the total annotation cost of $N_I+N_L$. The reduced cost is computed as: 
\begin{equation}
    Cost_{reduced} = 1-\frac{N_I^{LOF}+N_L^{LOF}}{N_I^{RD}+N_L^{RD}}. 
\end{equation}
It ranges in $(-\infty, 1]$; the higher, the better. 
Table~\ref{tab:initial_selection} summarizes the comparison results on dataset $DS_{low}4$. 
\begin{table}[h!]
    \centering
    \caption{Comparing the effectiveness of LOF-based to randomly-built initial set on dataset $DS_{low}4$. K:1.}
    \label{tab:initial_selection}
    \begin{tabularx}{\columnwidth}{lccccccc}
    \toprule
    \multirow{2}{*}{\textbf{Classifier}} & \multicolumn{3}{c}{\textbf{Random initialization}} &  \multicolumn{3}{c}{\textbf{LOF initialization}} & \multirow{2}{0.05\textwidth}{\textbf{Reduced cost}} \\
    \cmidrule{2-7}
    & \textbf{$N_{I}$} & \textbf{$N_L$} & \textbf{F1} & \textbf{$N_{I}$} & \textbf{$N_L$} & \textbf{F1} & \\
    \midrule
    KNN & 740 & 60 & 0.7719 & 400 & 212 & 0.7719 & 23.5\% \\
    XGBoost & 740 & 68 & 0.7753 & 100 & 251 & 0.7753 & 56.5\% \\
    CatBoost & 740 & 9 & 0.7719 & 100 & 73 & 0.7719 & 76.9\% \\
    LightGBM & 740 & 258 & 0.7699 & 400 & 261 & 0.7706 & 33.8\% \\    
    \bottomrule
    \end{tabularx}
\end{table}

The results suggest that our outlier detection-initialized \ac{AL} method significantly reduces the overall annotation cost to achieve comparable performances. \ac{CatBoost} and \ac{XGBoost} diminish the initial set size from 740 to 100 and yield 76.9\% and 56.5\% cost reductions, respectively, while achieving the same F1-scores compared with randomly initialized counterparts. Compared with \ac{CatBoost} and \ac{XGBoost}, \ac{KNN} and \ac{LightGBM} require more instances for initialization (400 rather than 100), possibly due to their low efficiency in learning imbalanced data. Yet, the \ac{LOF} initialization method manages to reduce their costs by 23.5\% and 33.8\%, respectively.

\section{Conclusion and Discussion}
This paper presents an \acf{ODEAL} framework to reduce the workload of \ac{QC} experts on ocean data quality assessment tasks. By exploiting the ability of \ac{AL} query strategies in selecting the most informative samples, our method achieved increases in F1-score of \ac{KNN}, \ac{XGBoost}, \ac{CatBoost}, and \ac{LightGBM} by 465.5\%, 46.8\%, 47.1\%, and 63.2\%, respectively with merely 100 queried samples. The \ac{LOF}-based initial set construction approach successfully identified 3 erroneous instances within top-100 ranked samples from a highly imbalanced dataset and accomplished a great decrease of annotation cost by 76.9\% for \ac{CatBoost}. 
To our knowledge, this is the first study dedicated to applying \ac{AL} to ocean data quality assessment. 
The promising experimental results on real Argo observatory data provide strong evidence of the effectiveness of this methodology. 

Nonetheless, there are some limitations associated with this work. First, classifier performances remain sub-optimal on highly imbalanced datasets, which calls for data-cleaning operations or more proficient learning models. Conflicts may exist in data distribution resulting from the inconsistency of data labeling over a long time period.  
Second, we exploit pool-based \ac{AL} methods, which require access to the entire dataset and might be unsuitable for real-time applications. Stream-based \ac{AL} that decides on the current instance for labeling could be more appropriate in such scenarios.

\section*{Acknowledgment}
We thank EuroArgo (Mr Thierry Carval and Mr Jean-Marie Baudet) and MARIS (Mr Peter Thijsse) for discussing the selected data sets, quality control processes, and data labels. This work has been partially funded by the European Union's Horizon research and innovation program by the  CLARIFY (860627), BLUECLOUD 2026 (101094227), ENVRI-FAIR (824068) and ARTICONF (825134), by the LifeWatch ERIC, and by the NWO LTER-LIFE project.

\bibliographystyle{IEEEtran}
\bibliography{references}

\begin{thebibliography}{10}
\providecommand{\url}[1]{#1}
\csname url@samestyle\endcsname
\providecommand{\newblock}{\relax}
\providecommand{\bibinfo}[2]{#2}
\providecommand{\BIBentrySTDinterwordspacing}{\spaceskip=0pt\relax}
\providecommand{\BIBentryALTinterwordstretchfactor}{4}
\providecommand{\BIBentryALTinterwordspacing}{\spaceskip=\fontdimen2\font plus
\BIBentryALTinterwordstretchfactor\fontdimen3\font minus \fontdimen4\font\relax}
\providecommand{\BIBforeignlanguage}[2]{{%
\expandafter\ifx\csname l@#1\endcsname\relax
\typeout{** WARNING: IEEEtran.bst: No hyphenation pattern has been}%
\typeout{** loaded for the language `#1'. Using the pattern for}%
\typeout{** the default language instead.}%
\else
\language=\csname l@#1\endcsname
\fi
#2}}
\providecommand{\BIBdecl}{\relax}
\BIBdecl

\bibitem{argo2001argo}
\BIBentryALTinterwordspacing
D.~Roemmich, O.~Boebel, Y.~Desaubies, H.~Freeland, K.~Kim, B.~King, P.-Y. Le~Traon, R.~Molinari, B.~W. Owens, S.~Riser, U.~Send, K.~Takeuchi, and S.~Wijffels, ``Argo : The global array of profiling floats,'' \emph{Observing the Oceans in the 21st Century}, 2001. [Online]. Available: \url{https://archimer.ifremer.fr/doc/00090/20097/}
\BIBentrySTDinterwordspacing

\bibitem{merrifield2009global}
M.~Merrifield, T.~Aarup, A.~Allen, A.~Aman, P.~Caldwell, E.~Bradshaw, R.~Fernandes, H.~Hayashibara, F.~Hernandez, B.~Kilonsky \emph{et~al.}, ``The global sea level observing system (gloss),'' \emph{Proceedings of the OceanObs}, vol.~9, 2009.

\bibitem{favali2009emso}
P.~Favali and L.~Beranzoli, ``Emso: European multidisciplinary seafloor observatory,'' \emph{Nuclear Instruments and Methods in Physics Research Section A: Accelerators, Spectrometers, Detectors and Associated Equipment}, vol. 602, no.~1, pp. 21--27, 2009.

\bibitem{lin2020ocean}
M.~Lin and C.~Yang, ``Ocean observation technologies: A review,'' \emph{Chinese Journal of Mechanical Engineering}, vol.~33, no.~1, pp. 1--18, 2020.

\bibitem{cummings2011ocean}
J.~A. Cummings, ``Ocean data quality control,'' \emph{Operational oceanography in the 21st Century}, pp. 91--121, 2011.

\bibitem{abeysirigunawardena2015data}
D.~Abeysirigunawardena, M.~Jeffries, M.~G. Morley, A.~O. Bui, and M.~Hoeberechts, ``Data quality control and quality assurance practices for ocean networks canada observatories,'' in \emph{OCEANS 2015-MTS/IEEE Washington}.\hskip 1em plus 0.5em minus 0.4em\relax IEEE, 2015, pp. 1--8.

\bibitem{diamant2020cross}
R.~Diamant, I.~Shachar, Y.~Makovsky, B.~M. Ferreira, and N.~A. Cruz, ``Cross-sensor quality assurance for marine observatories,'' \emph{Remote Sensing}, vol.~12, no.~21, p. 3470, 2020.

\bibitem{skaalvik2023challenges}
A.~M. Sk{\aa}lvik, C.~Saetre, K.-E. Fr{\o}ysa, R.~N. Bj{\o}rk, and A.~Tengberg, ``Challenges, limitations, and measurement strategies to ensure data quality in deep-sea sensors,'' \emph{Frontiers in Marine Science}, vol.~10, p. 1152236, 2023.

\bibitem{zhou2018data}
Y.~Zhou, R.~Qin, H.~Xu, S.~Sadiq, and Y.~Yu, ``A data quality control method for seafloor observatories: The application of observed time series data in the east china sea,'' \emph{Sensors}, vol.~18, no.~8, p. 2628, 2018.

\bibitem{castelao2021machine}
G.~P. Castel{\~a}o, ``A machine learning approach to quality control oceanographic data,'' \emph{Computers \& Geosciences}, vol. 155, p. 104803, 2021.

\bibitem{mieruch2021salaciaml}
S.~Mieruch, S.~Demirel, S.~Simoncelli, R.~Schlitzer, and S.~Seitz, ``Salaciaml: A deep learning approach for supporting ocean data quality control,'' \emph{Frontiers in Marine Science}, vol.~8, p. 611742, 2021.

\bibitem{demirel2021deep}
S.~Demirel, S.~Mieruch, and R.~Schlitzer, ``Deep learning for supporting ocean data quality control,'' \emph{Bollettino di Geofisica}, vol.~12, p. 121, 2021.

\bibitem{xin2023robust}
R.~Xin, H.~Liu, P.~Chen, and Z.~Zhao, ``Robust and accurate performance anomaly detection and prediction for cloud applications: a novel ensemble learning-based framework,'' \emph{Journal of Cloud Computing}, vol.~12, no.~1, pp. 1--16, 2023.

\bibitem{settles2009active}
B.~Settles, ``Active learning literature survey,'' University of Wisconsin--Madison, Computer Sciences Technical Report 1648, 2009.

\bibitem{argo2000argo}
G.~Argo, ``Argo float data and metadata from global data assembly centre (argo gdac),'' \emph{Seanoe}, 2000.

\bibitem{liu2008isolation}
F.~T. Liu, K.~M. Ting, and Z.-H. Zhou, ``Isolation forest,'' in \emph{2008 eighth ieee international conference on data mining}.\hskip 1em plus 0.5em minus 0.4em\relax IEEE, 2008, pp. 413--422.

\bibitem{scholkopf1999support}
B.~Sch{\"o}lkopf, R.~C. Williamson, A.~Smola, J.~Shawe-Taylor, and J.~Platt, ``Support vector method for novelty detection,'' \emph{Advances in neural information processing systems}, vol.~12, 1999.

\bibitem{breunig2000lof}
M.~M. Breunig, H.-P. Kriegel, R.~T. Ng, and J.~Sander, ``Lof: identifying density-based local outliers,'' in \emph{Proceedings of the 2000 ACM SIGMOD international conference on Management of data}, 2000, pp. 93--104.

\bibitem{fix1989discriminatory}
E.~Fix and J.~L. Hodges, ``Discriminatory analysis. nonparametric discrimination: Consistency properties,'' \emph{International Statistical Review/Revue Internationale de Statistique}, vol.~57, no.~3, pp. 238--247, 1989.

\bibitem{chen2016xgboost}
T.~Chen and C.~Guestrin, ``Xgboost: A scalable tree boosting system,'' in \emph{Proceedings of the 22nd acm sigkdd international conference on knowledge discovery and data mining}, 2016, pp. 785--794.

\bibitem{prokhorenkova2018catboost}
L.~Prokhorenkova, G.~Gusev, A.~Vorobev, A.~V. Dorogush, and A.~Gulin, ``Catboost: unbiased boosting with categorical features,'' \emph{Advances in neural information processing systems}, vol.~31, 2018.

\bibitem{ke2017lightgbm}
G.~Ke, Q.~Meng, T.~Finley, T.~Wang, W.~Chen, W.~Ma, Q.~Ye, and T.-Y. Liu, ``Lightgbm: A highly efficient gradient boosting decision tree,'' \emph{Advances in neural information processing systems}, vol.~30, 2017.

\end{thebibliography}

\end{document}